\documentclass{article}
\let\preprintaddcontentsline\addcontentsline
\usepackage{iclr2027_conference,times}
\let\addcontentsline\preprintaddcontentsline
\usepackage[T1]{fontenc}
\usepackage[utf8]{inputenc}
\usepackage{microtype}
\usepackage{amsmath,amssymb}
\usepackage{graphicx,booktabs,tabularx,float}
\usepackage[table]{xcolor}
\usepackage{subcaption} 
\definecolor{line-blue}{RGB}{243,248,252}
\definecolor{mydarkblue}{rgb}{0,0.08,0.45}
\usepackage[colorlinks,allcolors=mydarkblue]{hyperref}
\usepackage{wrapfig} 

\floatstyle{ruled}
\newfloat{algorithm}{htbp}{loa}
\floatname{algorithm}{Algorithm}
\newcounter{algline}
\newlength{\algindent}
\newenvironment{pseudocode}{%
  \small\setlength{\algindent}{0pt}%
  \begin{list}{{\footnotesize\arabic{algline}:}}{%
    \usecounter{algline}\setlength{\leftmargin}{2.5em}%
    \setlength{\labelwidth}{1.8em}\setlength{\labelsep}{.5em}%
    \setlength{\topsep}{2pt}\setlength{\itemsep}{0pt}\setlength{\parsep}{0pt}}%
}{\end{list}}
\newcommand{\Require}{\item[]\textbf{Input:} }
\newcommand{\State}{\item\hspace*{\algindent}\ignorespaces}
\newcommand{\For}[1]{\State\textbf{for} #1 \textbf{do}\addtolength{\algindent}{1em}}
\newcommand{\If}[1]{\State\textbf{if} #1 \textbf{then}\addtolength{\algindent}{1em}}
\newcommand{\EndFor}{\addtolength{\algindent}{-1em}\State\textbf{end for}}
\newcommand{\EndIf}{\addtolength{\algindent}{-1em}\State\textbf{end if}}
\newcommand{\Else}{\addtolength{\algindent}{-1em}\State\textbf{else}\addtolength{\algindent}{1em}}
\newcommand{\ElsIf}[1]{\Else{} \textbf{if} #1 \textbf{then}}

\title{Progress-Aware On-Policy Distillation for Multimodal Agents}
\author{
  \begin{tabular}{@{}l@{}}
    \textbf{Meng Yang}\thanks{Work done during internships at Alibaba Cloud.} \quad \textbf{Rongxue Li}\protect\footnotemark[1] \quad \textbf{Shurui Li \quad Hao Zhou \quad Yongliang Tao}\\
    \textbf{Lulu Hu \quad Bin Yang \quad Zhao Xu \quad Weihua Luo \quad Bowen Xu}\thanks{Corresponding author: \texttt{yangmeng.ym@mail.ustc.edu.cn}, \texttt{bowen.xbw@alibaba-inc.com}} \\
     {\normalfont Alibaba Group}
  \end{tabular}\\
}
\hypersetup{pdftitle={Progress-Aware On-Policy Distillation for Multimodal Agents}, pdfauthor={}}

\iclrfinalcopy
\begin{document}
\begingroup
\renewcommand{\thefootnote}{\fnsymbol{footnote}}
\maketitle
\endgroup
\lhead{}
\renewcommand{\headrulewidth}{0pt}
\begin{abstract}
Standard on-policy distillation (OPD) applies a uniform objective across
student-visited states.
In long-horizon multimodal tasks, however, local teacher alignment can
become decoupled from realized student progress.
We identify \emph{low-progress continuation}, where compact agents
continue searching or invoking tools without meaningfully updating
their answer state.
We present \textbf{$\Delta$-OPD}, a progress-aware distillation method
that conditions teacher supervision on the student's realized progress.
We quantify progress through the magnitude of the step-wise change
in \emph{answer readiness}, which measures the student's support
for the reference answer at its current state.
$\Delta$-OPD retains reverse KL for progressing interactions and applies
forward KL to low-progress interactions to strengthen supervision on
teacher-supported alternatives.
For verified-success trajectories, it masks low-progress continuation
losses when readiness exceeds the trajectory-specific successful-answer
reference.
The method operates entirely on student-generated trajectories,
requires no additional rollouts, and introduces no inference-time overhead.
Experiments on OmniGAIA, VideoDR, and VisualSimpleQA show improved
average accuracy for both Qwen3.5-4B and Qwen3.5-2B students.
On OmniGAIA, $\Delta$-OPD improves Qwen3.5-4B accuracy from
\textbf{40.8\%} under standard OPD to \textbf{45.4\%}, while reducing
the budget-exhaustion rate from \textbf{15.3\%} to \textbf{9.8\%},
a relative reduction of \textbf{35.9\%}.
Ablation studies show that selecting where to apply forward KL matters
beyond its overall frequency, while readiness-based masking provides
further gains.
Code is provided in the supplementary material.
\end{abstract}

\section{Introduction}
\label{sec:intro}

Multimodal agents increasingly solve complex tasks by combining heterogeneous
perception, long-context reasoning, and interaction with external tools.
Recent omni-modal models and benchmarks extend this setting to tasks involving
images, video, audio, web retrieval, and multi-turn tool use
\citep{team2026qwen3,li2026omnigaia}.
Transferring these capabilities to compact models is particularly challenging
because each interaction changes the evidence available to the agent and,
consequently, the states it encounters next.

On-policy distillation (OPD) is well suited to this setting: the student generates
its own trajectories while a stronger teacher provides dense token-level
supervision at the states the student actually visits
\citep{agarwal2024policy}.
Recent work has nevertheless revealed that long-horizon OPD can suffer from
increasing interaction length, state-distribution drift, and uneven supervision
across turns
\citep{luo2026demystifying,wang2026tcod,li2026policy,
zhong2026sod,li2026rethinking}.

Standard OPD applies the same local distillation objective across student-visited
states, effectively treating local imitation as equally useful throughout a
trajectory.
In long-horizon multimodal interaction, however, we find that local alignment with
a stronger teacher can become decoupled from realized student progress.
The consequence is a characteristic failure mode that we call
\emph{low-progress continuation}.
Unlike length or repetition inflation, low-progress continuation concerns whether
interaction changes the student's answer state: a long trajectory may remain
productive, while a non-repetitive trajectory may already have stalled.
These continuations need not be locally implausible.
Another search or perception operation can remain well supported under the current
history while the trajectory itself has stopped making progress.
Repeatedly optimizing local agreement can therefore continue refining an
interaction mode that is reasonable in isolation but no longer useful to the
student.

A natural alternative is to modulate supervision by local teacher--student discrepancy. However, our behavioral analysis shows that local discrepancy is only weakly predictive of downstream interaction failure, whereas changes in the student's answer state provide a stronger signal (Section~\ref{sec:motivation_alignment}).
This suggests that local alignment and realized interaction progress capture different properties of a visited state. These observations lead to the central principle of our approach:
\emph{given the same teacher and the same student-visited state, supervision should
depend on what the interaction actually accomplished for the student, rather than on local agreement with the teacher alone}.
We characterize this effect through the student's \emph{answer readiness}: how strongly it would support the reference answer if required to answer at the current state. The change in readiness across consecutive states measures the realized effect of
an interaction. For successful trajectories, the readiness of the state from which the student produces a verified correct answer further provides a trajectory-specific reference for interpreting low-progress interaction.

We therefore introduce \textbf{$\Delta$-OPD}, which makes realized student
progress an explicit control signal for on-policy distillation.
Interactions that continue to change the student's answer state retain reverse KL,
refining behavior the student is already able to realize.
Low-progress interactions use forward KL to expand supervision toward
teacher-supported alternatives, unless the student's readiness already exceeds its
trajectory-specific successful-answer reference, in which case continuation
supervision is withheld.
For trajectories without a verified correct completion, no readiness-based
withholding is applied.
$\Delta$-OPD keeps training rollouts entirely student-generated, requires no
additional teacher rollout, and introduces no inference-time component.

We evaluate $\Delta$-OPD by distilling Qwen3.5 students on OmniGAIA.
For Qwen3.5-4B, $\Delta$-OPD improves task accuracy from
\textbf{40.8\%} under standard OPD to \textbf{45.4\%}, while reducing the
budget-exhaustion rate from \textbf{15.3\%} to \textbf{9.8\%},
a relative reduction of \textbf{35.9\%}.
Trajectory-level analyses further show that these gains are not explained by
merely shortening trajectories: $\Delta$-OPD spends less time in low-progress
interaction regions while preserving productive long-horizon behavior.
Across OmniGAIA and two out-of-distribution benchmarks, the gains also transfer
to a smaller Qwen3.5-2B student, indicating that the proposed progress-conditioned
distillation principle generalizes across student scales.

Our contributions are threefold:
\begin{itemize}
    \setlength{\itemsep}{2pt}
    \item
    We identify \emph{low-progress continuation} as a failure mode of long-horizon
    on-policy distillation, showing that local teacher alignment can become
    decoupled from realized student progress.

    \item
    We introduce $\Delta$-OPD, which uses realized progress and a
    trajectory-specific readiness reference to adapt teacher supervision across
    interaction states.

    \item
    Experiments on OmniGAIA and two out-of-distribution benchmarks show that
    $\Delta$-OPD improves task accuracy over standard OPD across student scales,
    while substantially reducing budget exhaustion and time spent in low-progress
    interaction.
\end{itemize}

\section{Behavioral Motivation: From Premature Answering to Low-Progress Continuation}
\label{sec:motivation}

Before introducing our method, we examine how learning from a stronger teacher
reshapes the interaction behavior of a compact multimodal agent.
We train a Qwen3.5-4B student using Qwen3.8-27B as the teacher, either by supervised
fine-tuning (SFT) on teacher trajectories or by standard on-policy distillation
(OPD) on student-generated trajectories, and compare the resulting behavior with
the teacher on OmniGAIA~\citep{li2026omnigaia}.

\subsection{SFT and OPD Exhibit Opposite Failure Modes}
\label{sec:motivation_behavior}

\paragraph{SFT inherits a stopping policy that the student's capability cannot support.}
SFT learns interaction behavior from teacher-generated trajectories, including when
the teacher chooses to stop.
The student does not simply memorize a fixed interaction length: although the
training trajectories contain an average of $3.6$ tool calls on easier questions,
the SFT student uses $7.8$ calls on harder test questions, closely matching the
teacher's $7.9$.
Its stopping behavior therefore adapts to task difficulty.

This behavioral alignment does not imply matched answer capability.
Among questions answered without any tool call, the teacher is correct on $48\%$,
whereas the SFT student is correct on only $33\%$.
Moreover, among questions that SFT answers incorrectly without using a tool, an OPD
student continues interacting and eventually solves $35\%$.
SFT can therefore inherit the shape of the teacher's stopping behavior without
acquiring the capability that makes the same stopping decisions reliable.
We refer to this behavior as \emph{premature answering}: the student commits to an
answer even though further interaction can still recover a correct solution.

\begin{figure}[t]
    \centering
    \begin{subfigure}[t]{0.32\linewidth}
        \centering
        \includegraphics[width=\linewidth]
        {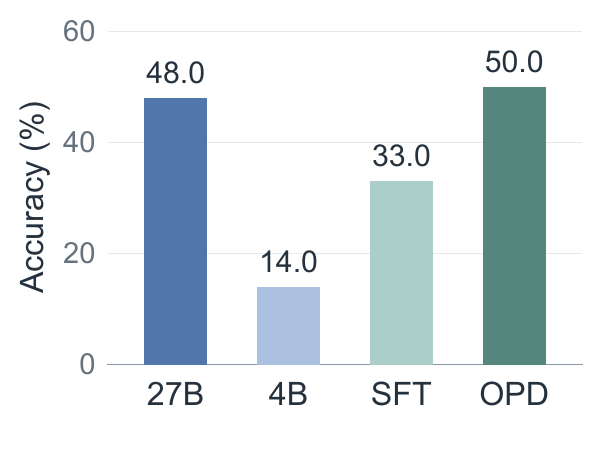}
        \caption{Zero-call accuracy}
        \label{fig:behavior_zero_call_accuracy}
    \end{subfigure}\hfill
    \begin{subfigure}[t]{0.32\linewidth}
        \centering
        \includegraphics[width=\linewidth]
        {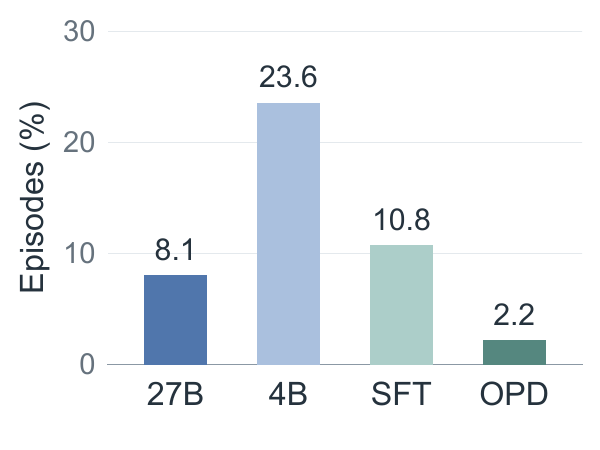}
        \caption{Zero-call rate}
        \label{fig:behavior_zero_call_rate}
    \end{subfigure}\hfill
    \begin{subfigure}[t]{0.32\linewidth}
        \centering
        \includegraphics[width=\linewidth]
        {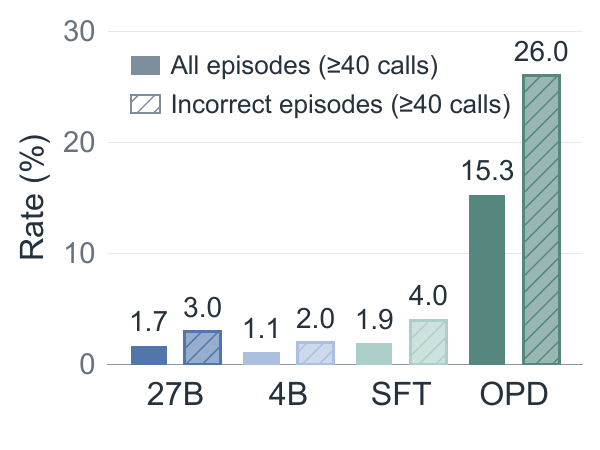}
        \caption{Long trajectories}
        \label{fig:behavior_long_trajectories}
    \end{subfigure}
    \caption{
    \textbf{SFT and OPD exhibit opposite failure modes.}
    We report zero-call accuracy, zero-call rate, and the fraction of abnormally
    long trajectories under SFT, standard OPD, and the teacher.
    SFT exhibits more premature answering, whereas OPD substantially reduces
    zero-call behavior but produces many more long trajectories.
    }
    \label{fig:behavior_failures}
    \vspace{-3mm}
\end{figure}

\paragraph{OPD corrects premature answering but overshoots into prolonged interaction.}
Standard OPD substantially corrects this failure.
The zero-call rate drops from $10.8\%$ under SFT to $2.2\%$ under OPD, while
zero-call accuracy rises from $33\%$ to $50\%$
(Figure~\ref{fig:behavior_failures}).
The behavioral shift, however, extends well beyond correcting premature answers.
Average interaction length increases from $7.8$ to $17.1$ calls, and the fraction
of trajectories exceeding $40$ calls rises from $1.9\%$ to $15.3\%$.
Only $17\%$ of web searches in these long trajectories are distinct, compared with
$86\%$ under SFT and the teacher.
Within these long trajectories, the additional interaction is therefore dominated
by redundant search rather than diverse evidence acquisition.

These statistics expose a behavioral tension rather than merely a preference for
short or long trajectories.
Standard OPD applies the same reverse-KL objective at every student-visited state.
This objective can refine a productive interaction mode, but provides no mechanism
for distinguishing it from a mode that the student keeps revisiting after its
interaction has ceased to change the answer state.
The issue is therefore not reverse KL in isolation, but its uniform application
across states with different realized effects.
This raises a more direct question: when the student continues to interact, is its
answer state actually still changing?

\subsection{Local Alignment Does Not Reveal Realized Progress}
\label{sec:motivation_alignment}

To understand whether continued interaction is actually productive, we inspect
the student's \emph{answer readiness}: how strongly the current state supports
the reference answer if the student were required to answer immediately.
We denote by $\Delta_t$ the readiness change induced by interaction $t$; its
formal definition is deferred to Section~\ref{sec:progress}.

Neither trajectory length nor surface repetition reliably identifies progress.
With the threshold later used by our method ($|\Delta_t|\leq0.05$), $39\%$ of
tool-calling turns exhibit little readiness change, yet $98\%$ of them are not
exact repetitions.
Conversely, even after eight interaction turns, $45\%$ of subsequent turns still
produce changes above the threshold.
Thus, long interaction can remain productive, while low-progress continuation
often occurs without explicit repetition.

Local teacher--student discrepancy does not resolve this ambiguity either.
Although it measures how far the student is from the teacher locally, it is much
less informative about downstream trajectory behavior than readiness change
(Appendix~\ref{app:progress_analysis}).
Moreover, trajectories observed at a low-progress state subsequently fail in
$72\%$ of cases and exhaust their interaction budget in $61\%$, compared with
$58\%$ and $42\%$ at progressing states.
These observations suggest that local teacher alignment and realized interaction
progress capture distinct properties.
They motivate conditioning distillation on the student's answer-readiness dynamics.
We formalize these quantities and develop $\Delta$-OPD in the next section.
\section{Method}
\label{sec:method}

\subsection{Answer Readiness and Realized Progress}
\label{sec:progress}

\begin{figure}[t]
    \centering
    \includegraphics[width=\linewidth]{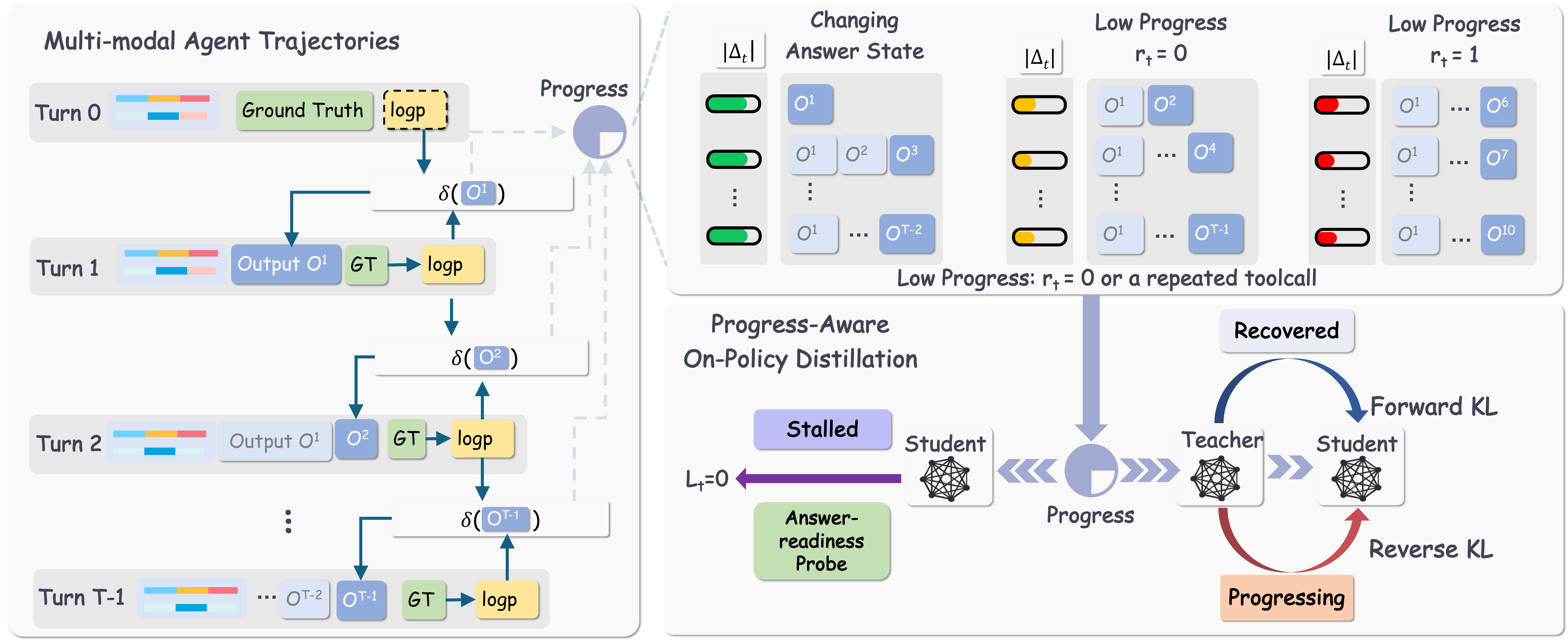}
    \caption{
    \textbf{Overview of $\Delta$-OPD.}
    Realized progress and a trajectory-specific readiness reference determine
    whether each student-generated interaction turn receives reverse KL,
    forward KL, or zero loss.
    Terminal completions retain reverse KL, and losses are averaged
    within and across turns.
    }
    \label{fig:method}
\end{figure}

Consider a student agent interacting with an environment over a multi-turn
trajectory
\begin{equation}
\xi=(h_0,a_0,o_0,\ldots,h_T,a_T),
\end{equation}
where $h_t$ denotes the interaction history before assistant turn $t$,
$a_t$ denotes the student's action at that turn, and $o_t$ is the resulting
environment observation.
We reserve $a_T$ for the terminal completion.
For each interaction turn $t<T$, the student produces $a_t$, the environment
returns $o_t$, and the resulting interaction is appended to the history to form
$h_{t+1}$.
In on-policy distillation (OPD), trajectories are generated entirely by the
student, while a stronger teacher provides token-level supervision at the
student-visited states.

Section~\ref{sec:motivation} shows that local teacher alignment does not reveal
whether an interaction actually advances the student.
We therefore characterize each visited state through the student's own answer
state and use its dynamics to determine how the corresponding interaction is
supervised.

\paragraph{Answer readiness.}
Let $y^*=(y_1^*,\ldots,y_m^*)$ be the reference answer of a training example.
For every student-visited state $h_t$, we define its \emph{answer readiness} as
the average log-likelihood that the current student assigns to the reference
answer if required to answer immediately:
\begin{equation}
V_t
=
\frac{1}{m}
\sum_{i=1}^{m}
\log
\bar{\pi}_\theta
\left(
y_i^*
\mid
h_t\oplus\texttt{<answer>},
y_{<i}^*
\right),
\label{eq:readiness}
\end{equation}
where $m$ is the number of reference-answer tokens and
$\bar{\pi}_\theta=\operatorname{sg}[\pi_\theta]$ denotes the current student
policy evaluated with gradients stopped.
The history $h_t$ is the actual interaction context seen by the agent, including
the multimodal task input, all preceding student turns, and the corresponding
tool observations.
For readiness scoring, we append the fixed \texttt{<answer>} prefix, condition on
the reference prefix $y^*_{<i}$, and score only the reference-answer tokens.

Thus, $V_t$ answers a counterfactual question:
\emph{if the student had to answer from the current state without further
interaction, how strongly would its predictive distribution support the correct
answer?}
Equivalently, $-V_t$ is the average per-token surprisal of the reference answer
under the current interaction context.
Larger $V_t$ therefore indicates stronger support for the correct answer.

\paragraph{Realized progress.}
We quantify the realized effect of interaction turn $t$ through the change in
answer readiness between its preceding and resulting states:
\begin{equation}
\Delta_t
=
V_{t+1}-V_t,
\qquad t<T.
\label{eq:progress}
\end{equation}

We use the magnitude of this change rather than its sign.
New evidence need not increase answer readiness monotonically: an informative
observation may invalidate the student's current hypothesis and temporarily
reduce the likelihood of the reference answer before subsequent reasoning
incorporates the new evidence.
Both positive and negative changes can therefore indicate a substantive update to
the student's answer state, whereas $|\Delta_t|\approx0$ indicates that the
interaction produced little change.

We refer to $|\Delta_t|$ as the measure of \emph{realized progress}.
Here, progress denotes substantive evolution of the student's answer state rather
than monotonic improvement in reference-answer likelihood.

\paragraph{Low-progress interaction.}
For each interaction turn $t<T$, we define
\begin{equation}
z_t
=
\mathbb{I}
\left[
|\Delta_t|\leq\tau
\right]
\lor
\mathbb{I}
\left[
\mathrm{Repeat}_t
\right],
\label{eq:stagnation}
\end{equation}
where $\tau$ is a global progress threshold, set to $0.05$ in the main
experiments, and $\mathrm{Repeat}_t=1$ when the tool call issued at turn $t$
exactly matches an earlier call in the same episode, including both the tool name
and its normalized arguments.

The two criteria capture complementary forms of low-progress interaction.
The repeat indicator identifies explicitly redundant tool use, whereas
$|\Delta_t|$ captures interactions that differ in surface form or action choice
but nevertheless leave the student's answer state nearly unchanged.
Examples include a reformulated search that retrieves essentially the same
evidence or a visual interaction that revisits content the student has already
inspected.
Edge cases, including truncated turns and states for which readiness cannot be
scored, are handled in Appendix~\ref{app:impl}.

Because $\Delta_t$ compares the states immediately before and after turn $t$, the
supervision assigned to that turn reflects its realized effect on the student's
answer state.
These quantities are computed after the student rollout is available and
therefore require no additional rollout or routing decision during generation.

\paragraph{Readiness reference.}
Low progress alone does not identify a unique supervision regime.
The behavioral analysis in Section~\ref{sec:motivation_behavior} shows that
teacher-shaped stopping behavior need not align with the student's own
capability.
We therefore anchor readiness to the student's own successful trajectory rather
than to a teacher-derived global criterion.

For a trajectory whose final completion $a_T$ is verified correct, let
\begin{equation}
V_{\mathrm{ans}}=V_T
\end{equation}
denote the readiness at the state from which that answer is produced.
For each interaction turn $t<T$, we define
\begin{equation}
r_t
=
\mathbb{I}
\left[
V_t>V_{\mathrm{ans}}
\right].
\label{eq:ready}
\end{equation}
A turn with $r_t=1$ starts from a state at which the student assigns higher
likelihood to the reference answer than at the state from which it successfully
answered.
The resulting reference is trajectory-specific and introduces no additional
globally tuned readiness threshold.

For trajectories without a verified correct completion, we set $r_t=0$ for all
interaction turns.
Together, $z_t$ and $r_t$ distinguish interactions that continue to change the
student's answer state from low-progress interactions, with the latter further
separated according to whether they exceed the student's successful-answer
readiness reference.

\subsection{\texorpdfstring{$\Delta$}{Δ}-OPD: Progress-Aware On-Policy Distillation}
\label{sec:delta_opd}

For token position $i$ of assistant turn $t$, let
\begin{equation}
q_{t,i}
=
\pi_\theta(\cdot\mid h_t,a_{t,<i}),
\qquad
p_{t,i}
=
\pi_{\mathrm{te}}(\cdot\mid h_t,a_{t,<i})
\end{equation}
denote the student and teacher next-token distributions under the same
student-generated autoregressive prefix.
We first express the routing rule using ideal full-vocabulary KL objectives;
training uses the top-$K$ implementations described in
Appendix~\ref{app:distillation_impl}, with $K=64$.

$\Delta$-OPD uses realized progress and answer readiness to determine the
supervision applied to each interaction turn:
\begin{equation}
\ell_{t,i}
=
\begin{cases}
D_{\mathrm{KL}}
\left(
q_{t,i}\,\|\,p_{t,i}
\right),
& z_t=0, \\[1mm]
D_{\mathrm{KL}}
\left(
p_{t,i}\,\|\,q_{t,i}
\right),
& z_t=1,\;r_t=0, \\[1mm]
0,
& z_t=1,\;r_t=1.
\end{cases}
\qquad t<T.
\label{eq:routing}
\end{equation}
The decision is made once per turn and applied to all valid supervised tokens of
that turn.

When $z_t=0$, the interaction materially changes the student's answer state.
We retain the reverse-KL objective of standard OPD, allowing the teacher to refine
behavior that the student is already able to realize.

When $z_t=1$ and $r_t=0$, the student's current interaction has produced little
realized progress but does not satisfy the readiness condition for withholding
further interaction supervision.
Forward KL transfers probability mass toward teacher-supported alternatives that
the student currently underweights, while leaving rollout generation unchanged.

The distinction can be seen from the ideal full-vocabulary token-level gradients.
Let $q$ and $p$ denote the student and teacher distributions at one token
position, and let $s_k$ be the student logit for token $k$:
\begin{align}
\frac{\partial D_{\mathrm{KL}}(q\|p)}
{\partial s_k}
&=
q_k
\left[
\log\frac{q_k}{p_k}
-
D_{\mathrm{KL}}(q\|p)
\right],
\label{eq:rkl_grad}\\
\frac{\partial D_{\mathrm{KL}}(p\|q)}
{\partial s_k}
&=
q_k-p_k.
\label{eq:fkl_grad}
\end{align}
For a teacher-preferred alternative with $q_k\ll p_k$, the reverse-KL correction
is attenuated by the student's own probability $q_k$, whereas the forward-KL
gradient remains approximately $-p_k$ even when the student assigns little
probability to that alternative.
Forward KL therefore preserves corrective pressure on precisely the
teacher-supported alternatives that the student's current interaction mode
underweights.

When $z_t=1$ and $r_t=1$, the interaction produces little realized progress from a
state whose readiness already exceeds the student's successful-answer reference.
We therefore assign that turn zero distillation loss rather than further
reinforcing its continuation.
We do not replace this signal with an explicit stopping target: readiness measures
support for the reference answer but does not certify what the student would
actually generate if it stopped.

$\Delta$-OPD does not insert teacher-generated actions into the training
trajectory or require a separate teacher continuation.
It changes only how the teacher distribution supervises student-generated turns
after the rollout has been collected.

\subsection{Training Objective}
\label{sec:objective}

$\Delta$-OPD applies the progress-conditioned distillation objective at the
interaction-turn level and uses turn-normalized loss aggregation.
Let $\mathcal{M}_t$ denote the set of valid assistant-token positions in turn
$t$ under the standard training mask.
For each turn, we first average the token-level distillation losses:
\begin{equation}
\mathcal{L}_t
=
\frac{1}{|\mathcal{M}_t|}
\sum_{i\in\mathcal{M}_t}
\ell_{t,i}.
\label{eq:turn_loss}
\end{equation}
For interaction turns $t<T$, the token-level loss $\ell_{t,i}$ is determined by
the routing rule in Eq.~\eqref{eq:routing}.

For the terminal completion, we retain the standard reverse-KL objective:
\begin{equation}
\ell_{T,i}
=
D_{\mathrm{KL}}
\left(
q_{T,i}\,\|\,p_{T,i}
\right).
\label{eq:terminal_loss}
\end{equation}

The trajectory-level objective then assigns equal weight to each assistant turn:
\begin{equation}
\mathcal{L}_{\Delta\text{-OPD}}
=
\frac{1}{T+1}
\sum_{t=0}^{T}
\mathcal{L}_t.
\label{eq:overall_loss}
\end{equation}
This turn-normalized reduction prevents long interaction turns from dominating
the optimization merely because they contain more tokens, while preserving the
progress-conditioned supervision assigned to each turn.

Both directions use multiple next-token candidates rather than the sampled token
alone, but the training loss does not sum over the entire vocabulary.
Forward KL sums over the teacher's top-$64$ tokens using full-vocabulary
probabilities, without renormalization within that support.
The reverse-KL branch uses the student's top-$64$ tokens and a policy-gradient
surrogate: student candidate weights are normalized within the retained support,
while the log-probabilities in the teacher--student log-ratio remain normalized
over the full vocabulary.
These are direction-specific top-$K$ approximations, not KL divergences between
two distributions both renormalized on a shared truncated support.
Appendix~\ref{app:distillation_impl} gives the support selection, normalization,
and surrogate objective, including the corresponding truncated forward-KL gradient.
Answer readiness and the resulting supervision decisions are used only during
training.
At inference time, neither the reference answer nor readiness computation is
required, and the trained student is deployed in the same manner as a standard
OPD-trained policy.
Implementation details, including masking and turn-level aggregation, are
provided in Appendix~\ref{app:impl}.
Figure~\ref{fig:method} summarizes how $\Delta$-OPD uses realized progress
and a trajectory-specific readiness reference to determine turn-level
distillation supervision.

\section{Experiments}
\label{sec:experiments}

\subsection{Experimental Setup}
\label{sec:experimental_setup}

\paragraph{Data and Benchmarks.}
We train on 2,156 tasks released with \textsc{OmniGAIA}~\citep{li2026omnigaia}.
For OPD-based methods, students generate trajectories on-policy from the original
queries and multimodal inputs, while SFT uses correctness-filtered trajectories
generated by Qwen3.8-27B.
We evaluate on \textsc{OmniGAIA} and two out-of-distribution benchmarks,
\textsc{VideoDR}~\citep{liu2026watching} and
\textsc{VisualSimpleQA}~\citep{wang2025visualsimpleqa}, covering omni-modal,
video, and image-based agent tasks.

\paragraph{Models and Baselines.}
Our primary setting uses Qwen3.8-27B~\citep{qwen38} as the teacher and
Qwen3.5-4B~\citep{qwen35blog} as the student.
We additionally evaluate Qwen3.5-2B under the same teacher to examine
generalization across student scales.
Both students use released post-trained checkpoints with native tool-use support;
thinking mode is disabled throughout training and evaluation.
We compare $\Delta$-OPD with the base student, SFT, standard
OPD~\citep{agarwal2024policy}, and recent multi-turn OPD methods, including
TCOD~\citep{wang2026tcod}, SOD~\citep{zhong2026sod}, and
TurnOPD~\citep{zhou2026turnopd}.

\paragraph{Evaluation Protocol.}
All methods use matched training and interaction settings whenever applicable and
are evaluated with identical prompts, decoding configurations, tool access, and
interaction budgets.
We report mean accuracy over four independent runs for each benchmark, together
with the unweighted average across the three benchmarks.
Full experimental and evaluation details are provided in
Appendix~\ref{app:experimental_details}.

\begin{table*}[t]
\centering
\small
\setlength{\tabcolsep}{3.0pt}

\resizebox{\linewidth}{!}{%
\begin{tabular}{lcccccccccc}
\toprule

\raisebox{-.65\baselineskip}{\textbf{Model / Method}}
& \multicolumn{9}{c}{\textbf{Category-Wise Breakdown}}
& \raisebox{-.65\baselineskip}{\textbf{Overall}} \\
\cmidrule(lr){2-10}

& \textbf{Geo.}
& \textbf{Tech.}
& \textbf{Hist.}
& \textbf{Fin.}
& \textbf{Sport}
& \textbf{Art}
& \textbf{Movie}
& \textbf{Sci.}
& \textbf{Food}
& \\

\midrule
\rowcolor{line-blue}
\multicolumn{11}{c}{\textbf{Proprietary Models}}\\

Gemini-3-Pro
& 65.2 & 59.2 & 62.1 & 72.0 & 78.4
& 52.8 & 48.5 & 42.3 & 88.9 & 62.5 \\

\midrule
\rowcolor{line-blue}
\multicolumn{11}{c}{\textbf{Open-source General Models}}\\

Qwen2.5-Omni-3B
& 0.0 & 2.0 & 4.5 & 0.0 & 0.0
& 0.0 & 0.0 & 3.9 & 0.0 & 1.4 \\

Qwen2.5-Omni-7B
& 1.5 & 4.1 & 7.5 & 4.0 & 0.0
& 2.8 & 0.0 & 7.7 & 5.6 & 3.6 \\

Qwen3-Omni-30B-A3B
& 8.7 & 14.3 & 11.9 & 28.0 & 10.8
& 13.9 & 9.1 & 15.4 & 22.2 & 13.3 \\

\midrule
\rowcolor{line-blue}
\multicolumn{11}{c}{\textbf{Open-source Agentic Models}}\\

OmniAtlas-Qwen2.5-7B
& 8.7 & 18.4 & 16.4 & 4.0 & 16.2
& 22.2 & 3.0 & 7.7 & 22.2 & 13.3 \\

OmniAtlas-Qwen3-30B-A3B
& 10.1 & 30.6 & 29.9 & 32.0 & 18.9
& 16.7 & 12.1 & 11.5 & 27.8 & 20.8 \\

Orchestra-o1-8B
& 21.7 & 32.7 & 37.9 & 12.0 & 29.7
& 16.7 & 45.5 & 38.5 & 38.9 & 30.0 \\

\midrule
\rowcolor{line-blue}
\multicolumn{11}{c}{\textbf{Qwen3.5-4B Training Methods}}\\

Qwen3.5-4B
& 26.1 & 38.8 & 35.9 & 44.0 & 32.4
& 38.0 & 23.2 & 29.5 & 29.6 & 33.0 \\

Qwen3.5-4B + SFT
& 29.7 & 41.8 & 48.5 & 52.0 & 37.8
& 30.6 & \textbf{42.4} & 28.8 & 50.0 & 39.6 \\

Qwen3.5-4B + OPD
& 31.9 & 41.8 & 51.5 & 44.0 & \textbf{41.9}
& 36.1 & 36.4 & 34.6 & \textbf{52.8} & 40.8 \\

\rowcolor{line-blue}
\textbf{Qwen3.5-4B + $\Delta$-OPD (Ours)}
& \textbf{36.2}
& \textbf{52.0}
& \textbf{53.8}
& \textbf{56.0}
& 39.2
& \textbf{44.4}
& 36.4
& \textbf{46.2}
& 44.4
& \textbf{45.4} \\

\bottomrule
\end{tabular}%
}

\vspace{-2pt}
\caption{
\textbf{Category-wise performance on OmniGAIA.}
Mean accuracy (\%) over four runs across nine domains.
Prior-model results are reproduced from previously reported evaluations;
bold indicates the best Qwen3.5-4B training result in each column.
}
\label{tab:main_results_domains}
\end{table*}

\subsection{Main Results}
\label{sec:main_results}

Table~\ref{tab:main_results_domains} reports category-wise performance on
\textsc{OmniGAIA}.
$\Delta$-OPD reaches \textbf{45.4\%} overall, improving standard OPD by
$4.6$ points ($40.8\%\rightarrow45.4\%$), SFT by $5.8$ points, and the
original student by $12.4$ points.
The gain is broad: $\Delta$-OPD obtains the best result among Qwen3.5-4B
training methods in six of nine domains, with particularly large improvements
over OPD in Finance ($+12.0$), Science ($+11.6$), and Technology ($+10.2$).

The resulting 4B student also exceeds previously reported open-source agentic
models, reaching 45.4\% versus 30.0\% for Orchestra-o1-8B and 20.8\% for
OmniAtlas-Qwen3-30B-A3B.
Since these results follow their originally reported evaluation settings, we
treat them as reference points rather than controlled comparisons.

\begin{table*}[t]
\centering
\caption{
\textbf{Cross-benchmark comparison across student scales.}
Mean task accuracy (\%) over four runs.
\textbf{Overall} is the full-benchmark score and \textbf{Avg.} is the unweighted
mean of the three Overall scores.
For VisualSimpleQA, \textbf{Hard} is the official 129-question subset and
\textbf{Normal} contains the remaining 371 questions.
Bold indicates the best training result within each student setting.
}
\label{tab:main_cross_setting}

\setlength{\tabcolsep}{3.2pt}
\small

\resizebox{\linewidth}{!}{%
\begin{tabular}{lcccccccccccc}
\toprule

\raisebox{-.65\baselineskip}{\textbf{Model / Method}}
& \multicolumn{4}{c}{\textbf{OmniGAIA}}
& \multicolumn{4}{c}{\textbf{VideoDR}}
& \multicolumn{3}{c}{\textbf{VisualSimpleQA}}
& \raisebox{-.65\baselineskip}{\textbf{Avg.}} \\

\cmidrule(lr){2-5}
\cmidrule(lr){6-9}
\cmidrule(lr){10-12}

& \textbf{Easy}
& \textbf{Medium}
& \textbf{Hard}
& \textbf{Overall}
& \textbf{Easy}
& \textbf{Medium}
& \textbf{Hard}
& \textbf{Overall}
& \textbf{Normal}
& \textbf{Hard}
& \textbf{Overall}
& \\

\midrule
\rowcolor[HTML]{F2F2F2}
\multicolumn{13}{c}{\textbf{Teacher Reference}} \\

Qwen3.8-27B
& 62.7 & 54.4 & 41.7 & 54.4
& 83.3 & 50.9 & 43.8 & 59.0
& 78.1 & 41.3 & 68.6
& 60.7 \\

\midrule
\rowcolor[HTML]{FAF8FF}
\multicolumn{13}{c}{\textit{Student: Qwen3.5-4B}} \\

Qwen3.5-4B
& 45.6 & 29.6 & 20.1 & 33.0
& 50.0 & 34.3 & 15.6 & 33.3
& 65.4 & 40.3 & 58.9
& 41.7 \\

\quad OPD
& 55.7 & 36.2 & 26.9 & 40.8
& 73.4 & 31.9 & 30.5 & 44.8
& 73.0 & \textbf{48.1} & 66.6
& 50.7 \\

\quad TCOD
& 58.6 & 40.3 & 25.6 & 43.3
& 70.3 & \textbf{43.8} & 26.6 & 46.8
& 75.5 & 42.6 & 67.0
& 52.4 \\

\quad SOD
& 56.6 & 38.1 & \textbf{31.4} & 42.9
& 67.2 & 37.5 & 32.8 & 45.5
& 77.4 & 40.3 & 67.8
& 52.1 \\

\quad TurnOPD
& 56.1 & 39.1 & 30.8 & 43.1
& 70.3 & 26.4 & 33.6 & 42.8
& 74.9 & 43.4 & 66.8
& 50.9 \\

\rowcolor[HTML]{EEF5FF}
\quad \textbf{$\Delta$-OPD (Ours)}
& \textbf{59.4} & \textbf{43.1} & 28.2 & \textbf{45.4}
& \textbf{75.8} & 42.4 & \textbf{38.3} & \textbf{51.8}
& \textbf{77.6} & 45.0 & \textbf{69.2}
& \textbf{55.5} \\

\midrule
\rowcolor[HTML]{FAF8FF}
\multicolumn{13}{c}{\textit{Student: Qwen3.5-2B}} \\

Qwen3.5-2B
& 19.7 & 14.7 & 9.0 & 15.1
& 43.8 & 26.4 & 21.9 & 30.5
& 52.2 & 22.9 & 44.6
& 30.1 \\

\quad OPD
& 33.6 & 18.4 & 9.6 & 21.7
& 54.7 & 27.1 & \textbf{23.4} & 34.8
& 64.4 & \textbf{34.1} & 56.6
& 37.7 \\

\quad TCOD
& 32.8 & 18.1 & 10.3 & 21.4
& 53.9 & 20.8 & 7.8 & 27.3
& 62.8 & 31.8 & 54.8
& 34.5 \\

\quad SOD
& \textbf{36.1} & 20.0 & 9.0 & 23.1
& 57.8 & \textbf{34.7} & 14.8 & \textbf{35.8}
& 59.0 & 32.6 & 52.2
& 37.0 \\

\quad TurnOPD
& 32.4 & 15.9 & 9.0 & 20.0
& \textbf{64.1} & 23.6 & 16.4 & 34.3
& 65.0 & 30.2 & 56.0
& 36.8 \\

\rowcolor[HTML]{EEF5FF}
\quad \textbf{$\Delta$-OPD (Ours)}
& 34.4 & \textbf{23.1} & \textbf{14.1} & \textbf{25.0}
& 57.8 & 31.9 & 10.2 & 33.3
& \textbf{66.8} & \textbf{34.1} & \textbf{58.4}
& \textbf{38.9} \\

\bottomrule
\end{tabular}%
}
\end{table*}

Table~\ref{tab:main_cross_setting} shows that the gains extend beyond OmniGAIA.
For the 4B student, $\Delta$-OPD achieves the best Overall score on all three
benchmarks, improving OPD by $4.6$, $7.0$, and $2.6$ points on OmniGAIA,
VideoDR, and VisualSimpleQA, respectively.
Its cross-benchmark average reaches \textbf{55.5\%}, exceeding OPD by $4.8$
points and the strongest competing distillation baseline by $3.1$ points.

The advantage largely persists at 2B.
$\Delta$-OPD improves OPD from $21.7\%$ to $25.0\%$ on OmniGAIA and from
$56.6\%$ to $58.4\%$ on VisualSimpleQA, yielding the highest cross-benchmark
average of \textbf{38.9\%}.
VideoDR is the exception, where SOD performs best (35.8\% versus 33.3\%).   $\Delta$-OPD also reduces the budget-exhaustion rate on OmniGAIA from $15.3\%$
   under standard OPD to $9.8\%$ (metric defined in Appendix~\ref{app:training_eval}).
Overall, the results show that progress-conditioned distillation transfers across
benchmark distributions and student scales, although the advantage is not
uniform across every individual setting.

\subsection{Ablation Study and Analysis}
\label{sec:ablation}

\begin{table*}[t]
\centering
\caption{
\textbf{Ablation study of $\Delta$-OPD.}
All variants use Qwen3.8-27B as the teacher and Qwen3.5-4B as the student.
Mean task accuracy (\%) is reported over four runs;
\textbf{Avg.} is the unweighted average across the three benchmarks.
Random FKL preserves the original FKL proportion within each trajectory while
randomizing the turns to which it is assigned.
}
\label{tab:ablation_components}

\setlength{\tabcolsep}{6pt}
\small
\begin{tabular*}{\textwidth}{@{\extracolsep{\fill}}lcccc@{}}
\toprule
\textbf{Method}
& \textbf{OmniGAIA}
& \textbf{VideoDR}
& \textbf{VisualSimpleQA}
& \textbf{Avg.} \\
\midrule
\textbf{$\Delta$-OPD (Full)}
& \textbf{45.4} & \textbf{51.8} & \textbf{69.2} & \textbf{55.5} \\
\midrule
\quad w/o FKL (replaced with zero loss)
& 43.0 & 48.2 & 67.2 & 52.8 \\
\quad w/o Progress-Based Routing (Random FKL)
& 42.3 & 47.3 & 66.7 & 52.1 \\
\quad w/o Readiness-Based Masking
& 44.1 & 50.1 & 68.3 & 54.2 \\
\quad w/o Turn-Level Normalization
& 44.7 & 50.9 & 68.6 & 54.7 \\
\bottomrule
\end{tabular*}
\end{table*}
Table~\ref{tab:ablation_components} isolates the contribution of each component.
Replacing FKL with zero loss reduces average accuracy from $55.5\%$ to $52.8\%$.
Randomly assigning FKL at the same within-trajectory frequency reduces it to
$52.1\%$. The benefit therefore depends on where FKL is applied, beyond its
overall frequency.
Readiness-based masking and turn-level normalization contribute $1.3$ and $0.8$
points, respectively, supporting selective removal of low-progress continuation
losses and balanced supervision across turns.

\subsection{Interaction Analysis}
\label{sec:interaction_analysis}

\begin{table}[!t]
\centering
\small
\setlength{\tabcolsep}{4pt}
\renewcommand{\arraystretch}{1.05}
\caption{\textbf{Tool-turn breakdown for $\Delta$-OPD and OPD on OmniGAIA.}
Counts cover the same 358 tasks, with one archived Qwen3.5-4B rollout per
task and policy and a 50-turn cap. OPD denotes standard on-policy
distillation, and $\Delta$-OPD uses the default threshold $\tau=0.05$.
A tool turn is labeled low-progress ($z_t=1$) if $|\Delta_t|\leq 0.05$
or it repeats a prior tool call. Early and late columns count low-progress
turns within the first five turns and after turn five, respectively.
Productive ($z_t=0$) denotes scored turns that satisfy neither condition.
Percentages use all tool turns for the corresponding policy as the denominator.}
\label{tab:interaction_summary}
\begin{tabularx}{\linewidth}{@{}Xrrrrrrr@{}}
\toprule
& & \multicolumn{4}{c}{\textbf{Low-progress} ($z_t=1$)}
  & \multicolumn{2}{c}{\textbf{Productive} ($z_t=0$)} \\
\cmidrule(lr){3-6}\cmidrule(l){7-8}
\textbf{Policy}
& \shortstack{\textbf{All tool}\\\textbf{turns}}
& \textbf{Count} & \textbf{\% of all} & \textbf{Early} & \textbf{Late}
& \textbf{Count} & \textbf{\% of all} \\
\midrule
$\boldsymbol{\Delta}$\textbf{-OPD}
& 4,145 & 2,611 & 63.0\% & 398 & 2,213 & 1,534 & 37.0\% \\
\textbf{OPD}
& 5,340 & 3,995 & 74.8\% & 478 & 3,517 & 1,345 & 25.2\% \\
\bottomrule
\end{tabularx}
\end{table}

\begin{table}[!t]
\centering
\small
\setlength{\tabcolsep}{4pt}
\renewcommand{\arraystretch}{1.05}
\caption{\textbf{OmniGAIA accuracy of $\Delta$-OPD, standard OPD, and
training variants.} The default $\Delta$-OPD uses $\tau=0.05$.
Uniform FKL and Uniform RKL apply forward and reverse KL, respectively,
at every turn. Fixed FKL/RKL preserves the per-trajectory FKL ratio of
$\Delta$-OPD and randomizes its placement across turns. Signed $\Delta$
replaces $|\Delta_t|\leq 0.05$ with $\Delta_t\leq 0.05$ in the progress
gate while retaining the repetition condition. The final two columns
vary the training threshold $\tau$.}
\label{tab:omnigaia_variant_accuracy}
\begin{tabularx}{\linewidth}{@{}Xcccccccc@{}}
\toprule
\textbf{Metric}
& $\boldsymbol{\Delta}$\textbf{-OPD}
& \textbf{OPD}
& \shortstack{\textbf{Uniform}\\\textbf{FKL}}
& \shortstack{\textbf{Uniform}\\\textbf{RKL}}
& \shortstack{\textbf{Fixed}\\\textbf{FKL/RKL}}
& \shortstack{\textbf{Signed}\\$\boldsymbol{\Delta}$}
& \shortstack{$\boldsymbol{\tau}$\\\textbf{0.025}}
& \shortstack{$\boldsymbol{\tau}$\\\textbf{0.1}} \\
\midrule
Accuracy
& 45.4\% & 40.8\% & 42.5\% & 43.3\% & 43.3\% & 44.0\% & 45.0\% & 43.1\% \\
\bottomrule
\end{tabularx}
\end{table}


\textbf{Quantitative analysis and human audit.}
Compared with OPD, $\Delta$-OPD reduces late low-progress turns from
3,517 to 2,213 ($37.1\%$), with turns after turn five accounting for
$94.2\%$ of the total reduction in low-progress turns
(Table~\ref{tab:interaction_summary}). Accuracy increases from
$40.8\%$ to $45.4\%$, compared with $42.5\%$, $43.3\%$, and $44.0\%$
for Uniform FKL, Uniform RKL, and Signed $\Delta$, respectively
(Table~\ref{tab:omnigaia_variant_accuracy}). Shuffling KL assignments while preserving the per-trajectory FKL ratio yields $43.3\%$,
supporting the contribution of state-dependent KL placement.
To assess whether small readiness changes reflect semantic
stagnation, we uniformly sample 100 interaction windows satisfying
$|\Delta_t|<0.05$ 
\begin{wrapfigure}{r}{0.48\columnwidth}
    \centering
    \includegraphics[width=\linewidth]{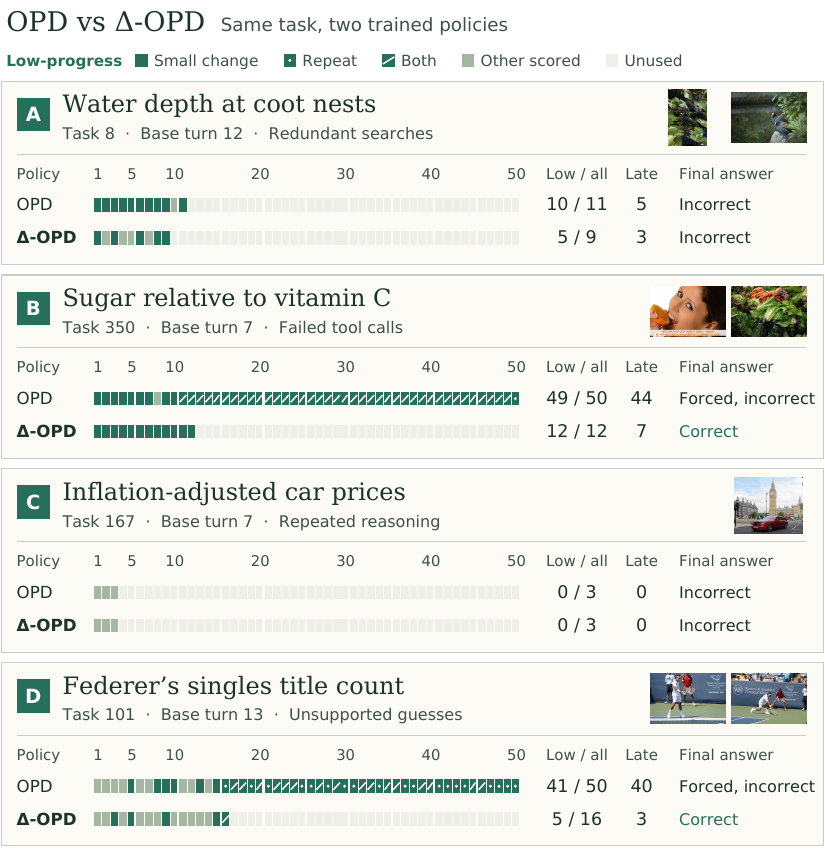}
    \caption{\textbf{Qualitative comparison of OPD and $\Delta$-OPD.}
    Each panel shows one archived Qwen3.5-4B rollout per policy.
    Squares denote tool turns in rollout order.
    A turn is labeled low-progress if $|\Delta_t|\leq0.05$ or it repeats a previous call with the same tool and arguments.
    \emph{Late} counts low-progress turns after turn five;
    \emph{Forced} marks an answer requested at the 50-turn cap.
    Behavior labels and base-turn indices refer to a separate
    base-model trajectory.}
    \label{fig:qualitative-policy-replays}
\end{wrapfigure}
from 586 windows across 55 Qwen3.5-4B tasks.
Human annotators examine each window with its preceding history,
while readiness scores, routing decisions, ground-truth answers,
and future turns are withheld. Meaningful progress requires new
task-relevant evidence, hypothesis narrowing, useful computation,
or error recovery; query reformulation alone is insufficient.
Of the 100 windows, 80 are judged stalled and 20 show meaningful
progress. The stalled cases comprise redundant searches (54),
failed tool calls (14), repeated reasoning (7), and unsupported
guesses (5). Together, these results support small readiness
changes as a useful, though imperfect, proxy for semantic stagnation
across behaviors beyond exact repetition, and show that
progress-conditioned routing improves accuracy while reducing
late low-progress continuation.

\noindent\textbf{Qualitative analysis.}
Figure~\ref{fig:qualitative-policy-replays} illustrates both the gains
and limitations of $\Delta$-OPD. On the nutrition and tennis tasks
(B and D), OPD reaches the 50-turn cap and answers incorrectly,
whereas $\Delta$-OPD answers correctly after 12 and 16 tool turns,
respectively. Fewer low-progress turns do not necessarily imply
success: on the coot task (A), their count decreases from 10 to 5,
but both policies remain incorrect. On the car-price task (C), both
answer incorrectly after three tool turns, none flagged as low
progress. Conversely, all 12 tool turns in the successful
$\Delta$-OPD nutrition rollout are flagged. These cases show that
the low-progress label is a diagnostic of readiness change or
call repetition, not a direct measure of a turn's usefulness or
of final-answer correctness.

\section{Related Work}

\textbf{On-Policy Distillation.} Recent studies identify severe state drift as a primary bottleneck in vanilla OPD ~\citep{luo2026demystifying, wang2026tcod, li2026policy, zhong2026sod}. Early student mistakes compound rapidly, pushing subsequent trajectories outside the teacher's reliable support. This drift often manifests as repetition loops~\citep{luo2026demystifying}, exhausted search budgets~\citep{li2026policy}, and superlinear divergence accumulation \cite{wang2026tcod, zhong2026sod}. To mitigate this, existing works restrict long-horizon exploration via temporal curricula~\citep{wang2026tcod}, probabilistically inject teacher actions to recover familiar states~\citep{li2026policy}, or downweight later turns based on teacher-student divergence ~\citep{zhong2026sod}. Instead of relying on rigid curricula or external heuristics, we dynamically resolve stagnation by conditioning supervision strictly on the student's own progress.

\textbf{Knowledge Distillation for MLLMs.}
Recent vision--language OPD methods primarily focus on \emph{internalizing privileged evidence} or \emph{refining supervision signals}. To internalize privileged information into tool-free policies, recent works apply self-distillation conditioned on cropped views~\citep{yuan2026visionopd}, zoomed visual imaginations~\citep{cai2026imagineopd}), or verified coding-agent traces~\citep{li2026spatialopsd}. Complementarily, another line of work refines the distillation objective itself. \citet{yoon2026decomposed} decompose language priors from visual grounding to steer gradients toward visual evidence. Across long-horizon tool interactions, $\Delta$-OPD tracks changes in the student's answer readiness to selectively gate distillation losses.

\textbf{Trajectory-level Imitation and Progress-Aware Supervision.}
Trajectory-based instruction tuning~\citep{zeng2024agenttuning, xu2026d} trains agents on expert demonstrations, but suffers from distribution shift as the learner's own actions alter future states. DAgger~\citep{ross2011reduction} and Generalized Knowledge Distillation (GKD)~\citep{agarwal2024policy} resolve this by querying expert or teacher supervision directly on student-visited states. Complementarily, in agent RL, IGPO and TIPS track step-wise changes in answer likelihood as dense process rewards~\citep{wang2026information,xie2026tips}. $\Delta$-OPD bridges these paradigms: while on-policy rollouts ensure state coverage, they do not indicate whether uniform imitation at every visited state is productive. We therefore condition teacher supervision on the realized progress of each interaction.

\vspace{-1mm}

\vspace{-1mm}

\section{Conclusion}
\label{sec:conclusion}

In this work, we study how on-policy distillation can better support
long-horizon multimodal agents.
Our analysis highlights a limitation of uniform teacher supervision:
closer teacher alignment does not necessarily translate into productive
interaction.
We propose $\Delta$-OPD, which uses realized changes in answer readiness
and a trajectory-specific reference to assign reverse KL, forward KL,
or zero loss to individual interaction turns.
Together with turn-level normalization, this design adapts supervision
to the student's evolving answer state.
Experiments on OmniGAIA and two out-of-distribution benchmarks show
improved average accuracy across two student scales.
On OmniGAIA, $\Delta$-OPD improves Qwen3.5-4B from 40.8\% to 45.4\%
accuracy while reducing budget exhaustion by 35.9\% relative to standard OPD.
Ablations support the importance of progress-conditioned KL routing,
with additional gains from readiness-based masking and turn-level normalization.
These findings highlight the value of using interaction progress to guide
agent distillation.
Future work will explore progress estimation without reference answers
and evaluate transfer across model families and broader agent environments.

\bibliographystyle{iclr2027_conference}
\bibliography{main}

\clearpage
\appendix
\section{Additional Experimental Details}
\label{app:experimental_details}

\subsection{Training Data and Benchmarks}
\label{app:data_benchmarks}

\paragraph{Training Data.}
We train on the 2,156 tasks released with
\textsc{OmniGAIA}~\citep{li2026omnigaia}.
For all OPD-based methods, the student starts from the original task query and
associated multimodal inputs and generates its interaction trajectories on-policy.

For the SFT baseline, we independently generate trajectories using Qwen3.8-27B
on the same training tasks and retain only those whose final answers are verified
as correct.
The resulting successful trajectories are used to fine-tune the Qwen3.5-4B
student.
SFT and OPD-based methods therefore share the same underlying training tasks,
while differing in both trajectory collection and training objective.

\paragraph{OmniGAIA.}
\textsc{OmniGAIA}~\citep{li2026omnigaia} is our primary evaluation benchmark.
It contains 360 omni-modal agent tasks requiring long-horizon reasoning over
images, audio, videos, and externally retrieved information.

\paragraph{VideoDR.}
\textsc{VideoDR}~\citep{liu2026watching} evaluates video-conditioned deep
research, requiring agents to identify informative visual evidence across video
frames and combine it with information retrieved from the open web.
We use the 100-question evaluation set and perform no benchmark-specific
training or adaptation.

\paragraph{VisualSimpleQA.}
\textsc{VisualSimpleQA}~\citep{wang2025visualsimpleqa} evaluates fact-seeking
question answering grounded in visual inputs.
We use the full 500-question set and additionally evaluate its 129-question hard
subset.

\subsection{Models}
\label{app:models}

Our primary setting uses Qwen3.8-27B~\citep{qwen38} as the teacher and
Qwen3.5-4B~\citep{qwen35blog} as the student.
We additionally evaluate Qwen3.8-27B$\rightarrow$Qwen3.5-2B to examine whether
the behavior of $\Delta$-OPD persists when the student capacity is reduced while
keeping the teacher fixed.
All Qwen3.5 students are used with thinking mode disabled during both training
and evaluation.

\subsection{Baseline Details}
\label{app:baselines}

Unless otherwise specified, all methods start from the same student checkpoint,
use the same training tasks, and interact with the same environment.

\paragraph{Base.}
The original Qwen3.5 student without additional task-specific post-training.

\paragraph{SFT.}
We perform supervised fine-tuning on correctness-filtered trajectories generated
by Qwen3.8-27B.
Unlike OPD-based methods, SFT trains on a fixed set of successful
teacher-generated trajectories rather than trajectories induced by the current
student policy.

\paragraph{OPD.}
We implement standard on-policy distillation following
\citet{agarwal2024policy}.
The student first generates its own trajectory, after which the teacher provides
token-level supervision at the states visited by the student.
The same reverse-KL objective is applied throughout the trajectory.

\paragraph{TCOD.}
TCOD~\citep{wang2026tcod} introduces a temporal curriculum for multi-turn
on-policy distillation by progressively adjusting the rollout horizon during
training.
Since our setting uses exclusively student-generated rollouts and does not
incorporate teacher-generated trajectories, we compare against its
forward-to-backward (F2B) variant under the same student--teacher and interaction
setting as our method.

\paragraph{SOD.}
SOD~\citep{zhong2026sod} performs step-wise on-policy distillation and modulates
teacher supervision according to the student--teacher discrepancy at individual
interaction steps.
We implement it under the same student-generated rollout setting as standard OPD
and set its step-wise grouping parameter to $n=4$.

\paragraph{TurnOPD.}
TurnOPD~\citep{zhou2026turnopd} makes on-policy distillation turn-aware through
adaptive rollout-depth control and turn-level loss normalization.
As no official implementation is publicly available, we reimplement TurnOPD
strictly following the method and training procedure described in the original
paper, within our common training framework.

\paragraph{$\Delta$-OPD.}
Our method follows the same student-generated rollout paradigm as standard OPD,
but conditions the distillation objective on the realized answer-readiness
dynamics of individual interactions.
Interactions that induce a meaningful change in the student's answer state use
reverse KL.
Low-progress interactions use forward KL unless their readiness exceeds the
trajectory-specific reference point, in which case their continuation
distillation loss is removed.
The terminal completion always retains the standard reverse-KL objective.
Detailed routing and implementation are provided in
Appendix~\ref{app:impl}.

\subsection{Training Details}
\label{app:training}

We implement all methods using the \texttt{verl}
framework~\citep{sheng2025hybridflow}.
Within each teacher--student configuration, online distillation methods share
the same student initialization, training task pool, interaction environment,
and available tools.
Table~\ref{tab:hparams} summarizes the default training configuration;
method-specific settings are described in Appendix~\ref{app:baselines}.
The distillation top-$k$ denotes the number of vocabulary tokens retained
when computing the KL loss.
No hyperparameters are tuned on the out-of-distribution benchmarks.

\begin{table}[t]
\centering
\caption{Default training hyperparameters for online distillation.
The progress threshold applies only to $\Delta$-OPD.}
\label{tab:hparams}
\begin{tabular}{lc}
\toprule
\textbf{Parameter} & \textbf{Value} \\
\midrule
Training Steps                  & 100 \\
Learning Rate                   & $1\times10^{-6}$ \\
Learning Rate Schedule          & Constant \\
Training Batch Size             & 64 \\
Rollouts per Prompt ($n$)        & 1 \\
Rollout Temperature              & 1.0 \\
Distillation Top-$k$             & 64 \\
Progress Threshold ($\tau$)      & 0.05 \\
\bottomrule
\end{tabular}
\end{table}

\subsection{Evaluation Details}
\label{app:training_eval}

For each trained model, we conduct four independent evaluation runs and sample
one interaction trajectory per query in each run.
We report the mean accuracy across the four runs.
The overall ``Avg.'' score is the unweighted arithmetic mean of the accuracies
on \textsc{OmniGAIA}, \textsc{VideoDR}, and \textsc{VisualSimpleQA}.

For \textsc{OmniGAIA}, we follow its official Pass@1 evaluation
protocol~\citep{li2026omnigaia}.
Exact answer matches are directly accepted, while non-exact predictions are
judged for semantic equivalence using
DeepSeek-V3.2~\citep{liu2025deepseek}.

Within each benchmark, all compared methods use identical prompts, decoding
settings, tool access, and maximum interaction budgets.
For analyses of interaction efficiency, we additionally report the average
number of tool calls and the budget-exhaustion rate, defined as the fraction of
episodes that exhaust the maximum interaction budget before producing a final
answer.

\section{Implementation Details}
\label{app:impl}

\subsection{Training Procedure}
\label{app:training_procedure}

Algorithm~\ref{alg:delta_opd} summarizes the training procedure of
$\Delta$-OPD.
Each training iteration first collects trajectories generated entirely by the
student.
After the rollout, we evaluate answer readiness at the visited states, compute
the realized progress of each interaction, and assign its distillation objective
according to Eqs.~\eqref{eq:stagnation}--\eqref{eq:routing}.
Token-level distillation losses are first averaged within each assistant turn,
and the resulting turn-level losses are then averaged across the trajectory.
No teacher-generated action is inserted into the training rollout.

\begin{algorithm}[htbp]
\caption{Training procedure of $\Delta$-OPD}
\label{alg:delta_opd}
\begin{pseudocode}
\Require student $\pi_\theta$, teacher $\pi_{\mathrm{te}}$, training batch with
reference answers, progress threshold $\tau$

\For{each task $(x,y^*)$ in the batch}
    \State Roll out the student trajectory
    $\xi=(h_0,a_0,o_0,\ldots,h_T,a_T)$
    \State Compute answer-readiness values
    $V_0,\ldots,V_T$ using Eq.~\eqref{eq:readiness}

    \If{$a_T$ is verified correct}
        \State $V_{\mathrm{ans}}\gets V_T$
    \EndIf

    \For{$t=0,\ldots,T-1$}
        \State $\Delta_t\gets V_{t+1}-V_t$
        \State
        $z_t\gets
        \mathbb{I}[|\Delta_t|\leq\tau]
        \lor
        \mathbb{I}[\mathrm{Repeat}_t]$

        \If{$a_T$ is verified correct}
            \State
            $r_t\gets
            \mathbb{I}[V_t>V_{\mathrm{ans}}]$
        \Else
            \State $r_t\gets 0$
        \EndIf

        \If{$z_t=0$}
            \State Assign reverse-KL loss to interaction turn $t$
        \ElsIf{$r_t=0$}
            \State Assign forward-KL loss to interaction turn $t$
        \Else
            \State Set the distillation loss of interaction turn $t$ to zero
        \EndIf
    \EndFor

    \State Assign standard reverse-KL loss to terminal completion $a_T$
    \State Average token losses within each turn
    \State Average turn-level losses over the trajectory
\EndFor

\State Average trajectory losses over the batch and update $\theta$
\end{pseudocode}
\end{algorithm}

\subsection{Answer-Readiness Computation}
\label{app:readiness_impl}

For each visited state $h_t$, answer readiness is evaluated using the same student
parameters that generated the trajectory and before the optimizer update for the
current batch.
The computation is performed with gradients stopped.

The state $h_t$ is the actual interaction context presented to the model at turn
$t$, including the multimodal task input, all preceding student turns, and the
corresponding tool observations.
For readiness scoring, we append the fixed \texttt{<answer>} prefix, condition on
the reference prefix $y^*_{<i}$, and evaluate the reference-answer tokens
autoregressively:
\begin{equation}
V_t
=
\frac{1}{m}
\sum_{i=1}^{m}
\log
\bar{\pi}_\theta
\left(
y_i^*
\mid
h_t\oplus\texttt{<answer>},
y_{<i}^*
\right),
\end{equation}
where
$\bar{\pi}_\theta=\operatorname{sg}[\pi_\theta]$.
Only the $m$ reference-answer tokens contribute to the readiness score, and the
same answer format is used at every visited state of an example.

This construction measures how strongly the student would support the reference
answer if required to answer directly from the current interaction state, without
additional reasoning or tool use.
Readiness is used only as a training-time diagnostic and does not modify the
student rollout.

\subsection{Interaction-Level Decisions}
\label{app:interaction_decisions}

For each interaction turn $t<T$, realized progress is computed from the change in
answer readiness between the states immediately before and after the interaction:
\begin{equation}
\Delta_t
=
V_{t+1}-V_t.
\end{equation}
The low-progress indicator combines the magnitude of this change with an
exact-repeat criterion:
\begin{equation}
z_t
=
\mathbb{I}[|\Delta_t|\leq\tau]
\lor
\mathbb{I}[\mathrm{Repeat}_t].
\end{equation}

$\mathrm{Repeat}_t$ is activated when the tool call issued at turn $t$ exactly
matches an earlier call in the same episode, including both the tool name and its
normalized arguments.
Thus, the repeat criterion captures explicitly redundant tool use, while the
readiness-change criterion can identify interactions that differ in surface form
but nevertheless leave the student's answer state nearly unchanged.

For trajectories with a verified correct terminal completion, we set
\begin{equation}
V_{\mathrm{ans}}=V_T
\end{equation}
and compute the trajectory-specific readiness indicator
\begin{equation}
r_t
=
\mathbb{I}[V_t>V_{\mathrm{ans}}],
\qquad t<T.
\end{equation}
For trajectories without a verified correct completion, we set $r_t=0$ for all
interaction turns, as in the main method.

Both $z_t$ and $r_t$ are computed only after the complete student trajectory has
been collected.
They therefore affect the distillation objective assigned to an observed
student-generated turn, rather than the rollout policy used to generate that turn.

\subsection{Token-Level Distillation and Turn Normalization}
\label{app:distillation_impl}

For each supervised token position $i$ of interaction turn $t$, the student and
teacher are evaluated under the same student-generated autoregressive prefix:
\begin{equation}
q_{t,i}
=
\pi_\theta(\cdot\mid h_t,a_{t,<i}),
\qquad
p_{t,i}
=
\pi_{\mathrm{te}}(\cdot\mid h_t,a_{t,<i}).
\end{equation}

For interaction turns $t<T$, the token-level objective follows
Eq.~\eqref{eq:routing}:
reverse KL is used when $z_t=0$, forward KL is used when
$z_t=1$ and $r_t=0$, and the distillation loss is set to zero when
$z_t=1$ and $r_t=1$.
All supervised tokens within the same interaction turn share the same
interaction-level decision.

The terminal completion $a_T$ does not receive an interaction-level progress
decision and retains the standard reverse-KL objective:
\begin{equation}
\ell_{T,i}
=
D_{\mathrm{KL}}
\left(
q_{T,i}\,\|\,p_{T,i}
\right).
\end{equation}

To prevent turns with more generated tokens from receiving disproportionately
large optimization weight, we use turn-normalized loss aggregation.
Let $\mathcal{M}_t$ denote the set of valid assistant-token positions in turn
$t$ under the training mask.
We first average the token-level losses within each turn:
\begin{equation}
\mathcal{L}_t
=
\frac{1}{|\mathcal{M}_t|}
\sum_{i\in\mathcal{M}_t}
\ell_{t,i}.
\label{eq:app_turn_loss}
\end{equation}
The trajectory loss is then computed by assigning equal weight to all assistant
turns, including the terminal completion:
\begin{equation}
\mathcal{L}_{\mathrm{traj}}
=
\frac{1}{T+1}
\sum_{t=0}^{T}
\mathcal{L}_t.
\label{eq:app_traj_loss}
\end{equation}
Turns routed to zero continuation supervision have
$\mathcal{L}_t=0$ and remain part of the turn-level reduction.
Finally, trajectory losses are averaged across examples in the training batch.

This reduction separates the choice of distillation direction from the amount of
text generated at a turn: the routing rule determines \emph{how} a state is
supervised, while turn normalization prevents long responses from dominating the
update solely because they contain more tokens.

Both KL directions use multiple next-token candidates
with direction-specific top-$K$ approximations,
rather than relying on the sampled token alone.
Answer-readiness computation, progress routing, and the teacher distribution are
required only during training; none introduces an inference-time component.

\subsection{Boundary Cases}
\label{app:edge_cases}

Incomplete interactions, truncated trajectories, and states for which answer
readiness cannot be evaluated are handled conservatively so that they do not
introduce spurious progress signals or readiness decisions.
Exact handling rules for these implementation-specific cases are provided with
the released code.

\section{Additional Analysis of Realized Progress}
\label{app:progress_analysis}

We provide additional evidence for the central observation in
Section~\ref{sec:motivation_alignment}: local teacher--student alignment does not
reveal whether the student's interaction is producing realized progress.
Our analysis covers $28{,}952$ tool-calling turns from $6{,}400$ on-policy
rollouts of the Qwen3.5-4B student over $2{,}139$ training questions, collected
during on-policy distillation with a turn budget of $16$.
Every visited state is annotated with its readiness change $\Delta_t$, the local
teacher--student discrepancy seen by OPD, and the outcome of the remaining
trajectory.

\paragraph{Length and repetition do not identify realized progress.}
Using the same threshold adopted by our method,
$|\Delta_t|\leq0.05$, $39\%$ of tool-calling turns fall into the low-progress
regime.
However, $98\%$ of these turns do not exactly repeat any earlier tool call, while
exact repetitions account for only $1.3\%$ of all tool-calling turns.
Thus, explicit repetition captures only a small fraction of interactions that
leave the student's answer state nearly unchanged.

Trajectory length is similarly insufficient.
Even after $8$ interaction turns, $45\%$ of subsequent turns still exhibit
readiness changes above the threshold.
Long trajectories therefore contain both productive interactions and
low-progress continuation, and interaction count alone cannot determine whether
the student's answer state is still changing.

\paragraph{Local discrepancy does not reveal realized progress.}
A natural alternative is the local teacher--student discrepancy already computed
during OPD.
We measure discrepancy as the per-turn mean of the token-level reverse KL
between student and teacher, evaluated under the same student-generated
autoregressive prefixes and the teacher top-$64$ support used by the OPD
objective.

Local discrepancy and realized progress capture fundamentally different
properties.
A student can disagree strongly with the teacher while still acquiring useful
evidence, or remain locally aligned while continued interaction produces little
change in its answer state.
We test this distinction using downstream outcomes that do not depend on the
low-progress criterion itself.
At every visited state, we evaluate how well local discrepancy and
$|\Delta_t|$ predict whether the remaining trajectory eventually fails,
exhausts its interaction budget, or enters exact repetition.

As shown in Table~\ref{tab:signal_predictiveness}, readiness dynamics are
consistently more informative than local teacher--student discrepancy.
For eventual task failure, $|\Delta_t|$ reaches an AUROC of $0.65$, compared
with only $0.47$ for local discrepancy.
For budget exhaustion, the corresponding values are $0.64$ and $0.46$.
For future exact repetition, readiness change reaches $0.56$, while local
discrepancy reaches $0.45$.
The below-chance performance of local discrepancy further shows that large local
student--teacher disagreement is not evidence of unproductive interaction; if
anything, it is slightly associated with states in which the student's behavior
is still changing.

\begin{table}[t]
    \centering
    \small
    \caption{
    \textbf{Local discrepancy does not reveal realized progress.}
    AUROC of per-state signals for downstream outcomes measured after the same
    visited state ($28{,}952$ tool-calling turns from $6{,}400$ rollouts).
    For local discrepancy, larger disagreement is scored toward failure; for
    $|\Delta_t|$, smaller readiness change is scored toward failure.
    }
    \label{tab:signal_predictiveness}
    \begin{tabular}{lccc}
        \toprule
        \textbf{Signal}
        & \textbf{Task Failure}
        & \textbf{Budget Exhaustion}
        & \textbf{Future Repetition} \\
        \midrule
        Local teacher--student discrepancy
        & 0.47
        & 0.46
        & 0.45 \\
        Readiness change $|\Delta_t|$
        & \textbf{0.65}
        & \textbf{0.64}
        & \textbf{0.56} \\
        \bottomrule
    \end{tabular}
    \vspace{-2mm}
\end{table}

\paragraph{Low-progress states predict undesirable continuation.}
The same separation is visible after discretizing interactions with the
readiness-change threshold.
A trajectory observed at a low-progress state subsequently fails in $72\%$ of
cases and exhausts its interaction budget in $61\%$.
For trajectories observed at a progressing state, the corresponding rates are
only $58\%$ and $42\%$.
Thus, low progress is not merely a geometric property of the student's
distribution: it is strongly associated with the downstream failure modes that
characterize inefficient long-horizon interaction.

Taken together, these results expose a missing variable in standard OPD.
Local teacher--student discrepancy tells us how the student differs from the
teacher at the current state, but not whether the student's own interaction is
still producing useful change.
Realized readiness dynamics reveal this distinction, motivating the
progress-conditioned supervision introduced by $\Delta$-OPD.

\clearpage

\section{Qualitative Examples of Distinct Low-Change Behaviors}
\label{app:behavioral-case-examples}

The following four hand-selected cases distinguish behaviors previously grouped
under a broad slow/stalled label. They are illustrative readings of archived
Qwen3.5-4B text, not human-validated labels or a new prevalence estimate.
Quoted passages are separate verbatim excerpts; task descriptions and
interpretations are summaries. Turn numbers count assistant responses, and
$\Delta_t$ is taken from the primary archived probe. All four focal turns
have no exact tool-call duplicate flag. Small score changes are reported for
context and do not determine the interpretation.

\subsection{Repeated Reasoning About an Already Resolved Result}
\label{app:case-p066}

P066 (task 167, turn 7; $\Delta_t=+0.028894$) concerns an inflation-adjusted
price ratio expressed as $1:X$. The preceding turn has already corrected the
ratio direction and numerically checked it. The focal response repeatedly
revisits this interpretation before returning the same requested result.

\begin{table}[H]
\centering
\small
\setlength{\tabcolsep}{4pt}
\renewcommand{\arraystretch}{1.12}
\caption{\textbf{Repeated reasoning without an exact repeated tool call (P066).}}
\label{tab:behavior-p066}
\begin{tabularx}{\linewidth}{@{}>{\raggedright\arraybackslash}p{0.23\linewidth}>{\raggedright\arraybackslash}X@{}}
\toprule
\textbf{Stage} & \textbf{Logged evidence} \\
\midrule
Prior tool result (turn 6) & ``Ratio (2015 : 1931) = 1 : 0.62'' \\
Focal assistant (turn 7) & ``Wait, I'm getting confused. Let me think about this more carefully.'' \\
Later in the same response & ``Actually, I think the issue is that I'm overthinking this.'' \\
Focal tool result & ``Final answer: 1 : 0.62'' \\
\bottomrule
\end{tabularx}
\end{table}

The evidence for a reasoning cycle is the repeated reconsideration of an
already checked result without a new required input or changed output.
Self-doubt alone would not establish stagnation, and differently written
tool arguments can still implement a redundant computation.

\subsection{Failed Retrieval During an Attempted Verification}
\label{app:case-p097}

P097 (task 350, turn 7; $\Delta_t=-0.004103$) seeks nutritional inputs for a
sugar-to-vitamin-C comparison. Earlier search snippets provide some values;
the student attempts to obtain more precise data from three USDA pages.

\begin{table}[H]
\centering
\small
\setlength{\tabcolsep}{4pt}
\renewcommand{\arraystretch}{1.12}
\caption{\textbf{A reasonable verification attempt blocked by retrieval errors (P097).}}
\label{tab:behavior-p097}
\begin{tabularx}{\linewidth}{@{}>{\raggedright\arraybackslash}p{0.23\linewidth}>{\raggedright\arraybackslash}X@{}}
\toprule
\textbf{Stage} & \textbf{Logged evidence} \\
\midrule
Focal assistant (turn 7) & ``Now let me gather more precise data by looking at specific USDA pages.'' \\
Requested pages & A single \texttt{page\_browser} call requests USDA FoodData Central records 16799, 11520, and 11517. \\
Returned errors & Each of the three requested URLs returns ``HTTP 404''; no nutritional table is returned. \\
\bottomrule
\end{tabularx}
\end{table}

This block produces no new nutritional input or completed calculation.
Its immediate limitation is retrieval failure, however, rather than a
demonstrated reasoning loop. Consulting an authoritative source is a reasonable
verification strategy; this single failed attempt does not establish that the
strategy itself is redundant or that the student is persistently stalled.

\clearpage
\subsection{An Unsupported Inference Used to Constrain Search}
\label{app:case-p027}

P027 (task 60, turn 10; $\Delta_t=+0.002644$) seeks the comic issue containing
a company's first appearance, needed for a date-difference calculation.
The student treats scattered mentions of Parker Industries and issue numbers
as support for a specific first-appearance claim.

\begin{table}[H]
\centering
\small
\setlength{\tabcolsep}{4pt}
\renewcommand{\arraystretch}{1.12}
\caption{\textbf{A specific claim without the required evidential link (P027).}}
\label{tab:behavior-p027}
\begin{tabularx}{\linewidth}{@{}>{\raggedright\arraybackslash}p{0.23\linewidth}>{\raggedright\arraybackslash}X@{}}
\toprule
\textbf{Stage} & \textbf{Logged evidence} \\
\midrule
Earlier query (turn 8) & \texttt{"Parker Industries" first created "Amazing Spider-Man" \#16 June 1963 Otto Octavius} \\
Focal assistant (turn 10) & ``This is referring to Amazing Spider-Man \#16.'' \\
Focal query & \texttt{"Amazing Spider-Man" \#16 June 1963 Parker Industries created} \\
One returned snippet & ``Information about the comic The Amazing Spider-Man \#16 [British], which has a publication date of September 1964 and was published by Marvel.'' \\
\bottomrule
\end{tabularx}
\end{table}

The preceding snippets do not establish that issue \#16 contains the company's
first appearance. The focal return above concerns an issue's publication date
and does not supply that missing link either. The observable problem is an
unsupported inference anchoring the search, not proof that every new hypothesis
is stagnant. This interpretation concerns the available record, rather than an
independent verification of comic-book history.

\subsection{Marginal Checking as a Boundary Case}
\label{app:case-p054}

P054 (task 115, turn 7; $\Delta_t=+0.025346$) needs a circus's first performance
date at a specified venue to compute a day difference. Turn 2 already retrieves
a schedule giving July 1--2; turn 7 retrieves a later report repeating those
dates and adding times of day.

\begin{table}[H]
\centering
\small
\setlength{\tabcolsep}{4pt}
\renewcommand{\arraystretch}{1.12}
\caption{\textbf{Repeated date information with additional corroboration (P054).}}
\label{tab:behavior-p054}
\begin{tabularx}{\linewidth}{@{}>{\raggedright\arraybackslash}p{0.23\linewidth}>{\raggedright\arraybackslash}X@{}}
\toprule
\textbf{Stage} & \textbf{Logged evidence} \\
\midrule
Prior search result (turn 2) & ``Lake Erie Speedway is excited to bring in the Carson Barnes Circus on Saturday, July 1st and Sunday, July 2nd.'' \\
Focal query (turn 7) & \texttt{Carson Barnes Circusaurus Show July 1 2017 Erie Pennsylvania first performance} \\
Focal search result & ``The Carson and Barns Circus will be at the Speedway Saturday and Sunday, July 1st and 2nd from 2pm until 5pm.'' \\
\bottomrule
\end{tabularx}
\end{table}

The old audit treated this as marginal checking because the required date was
already available and the added hours did not change the day-level calculation.
However, a later report can corroborate an earlier schedule. Its value depends
on whether uncertainty about that schedule remained unresolved; repeated
information alone is insufficient to rule out useful verification. We therefore
present this as a boundary case, rather than an unambiguous reasoning loop.

\end{document}